\pdfoutput=1

\documentclass[11pt]{article}

\usepackage[hyperref]{EMNLP2023}

\usepackage{times}
\usepackage{latexsym}
\usepackage{color}
\usepackage{graphicx}
\usepackage{tabularx}
\usepackage{longtable}
\usepackage[T1]{fontenc}
\usepackage{CJK}

\usepackage[utf8]{inputenc}

\usepackage{microtype}

\usepackage{inconsolata}
\usepackage{multirow}
\usepackage{booktabs}
\usepackage{amsmath}

\title{M3KE: A Massive Multi-Level Multi-Subject Knowledge Evaluation Benchmark for Chinese Large Language Models}

\author{
Chuang Liu\textsuperscript{\rm{1}}, 
Renren Jin\textsuperscript{\rm{1}}, 
Yuqi Ren\textsuperscript{\rm{1}}, 
Linhao Yu\textsuperscript{\rm{1}}, 
Tianyu Dong\textsuperscript{\rm{1}}, 
Xiaohan Peng\textsuperscript{\rm{1}},
Shuting Zhang\textsuperscript{\rm{1}}\\
\textbf{Jianxiang Peng}\textsuperscript{\rm{1}}\textbf{,} 
\textbf{Peiyi Zhang}\textsuperscript{\rm{1}}\textbf{,} 
\textbf{Qingqing Lyu}\textsuperscript{\rm{1}}\textbf{,} 
\textbf{Xiaowen Su}\textsuperscript{\rm{1}}\textbf{,}
\textbf{Qun Liu}\textsuperscript{\rm{2}} 
\and \textbf{Deyi Xiong}\textsuperscript{\rm{1}} \Thanks{~Corresponding author.} \\
\textsuperscript{1}
College of Intelligence and Computing, Tianjin University, Tianjin, China \\
\textsuperscript{2} Huawei Noah's Ark Lab, Hong Kong, China \\
\texttt{\{liuc\_09,rrjin,ryq20,linhaoyu,skydong112358,pengxiaohan\}@tju.edu.cn}\\
\texttt{\{shutingzhang,pjasonx,zhangpeiyi,qingq\_lv,suxiaowen,dyxiong\}@tju.edu.cn}\\
\texttt{qun.liu@huawei.com}
}

\begin{document}
\maketitle

\begin{abstract}
Large language models have recently made tremendous progress in a variety of aspects, e.g., cross-task generalization, instruction following. Comprehensively evaluating the capability of large language models in multiple tasks is of great importance. In this paper, we propose M3KE, a Massive Multi-Level Multi-Subject Knowledge Evaluation benchmark, which is developed to measure knowledge acquired by Chinese large language models by testing their multitask accuracy in zero- and few-shot settings. We have collected 20,477 questions from 71 tasks. Our selection covers all major levels of Chinese education system, ranging from the primary school to college, as well as a wide variety of subjects, including humanities, history, politics, law, education, psychology, science, technology, art and religion. All questions are multiple-choice questions with four options, hence guaranteeing a standardized and unified assessment process. We've assessed a number of state-of-the-art open-source Chinese large language models on the proposed benchmark. The size of these models varies from 335M to 130B parameters. Experiment results demonstrate that they perform significantly worse than GPT-3.5 that reaches an accuracy of $\sim$ 48$\%$ on M3KE. The dataset is available at \url{https://github.com/tjunlp-lab/M3KE}. 
\end{abstract}
\begin{CJK}{UTF8}{gbsn}
\section{Introduction}

Large Language Models (LLMs) \cite{Raffel-JMLR-2020-Exploring, Xue-NAACL-2021-mT5, Zhang-arxiv-2022-OPT, Brown-NeurIPS-2020-Language, Touvron-arxiv-2023-LLaMA, Scao-arxiv-2022-BLOOM, LLMSurvey, Zhou-arxiv-2023-A} have achieved remarkable progress in recent years, especially with the release of ChatGPT\footnote{https://openai.com/blog/chatgpt}, which is widely acknowledged to revolutionize the world of natural language processing and to transform AI and society \cite{OpenAI-blog-2023-Planning, Bubeck-arxiv-2023-Sparks,Huang-CoRR-2023, Cao-arxiv-2023-comprehensive}. Generally, LLMs are trained via self-supervised learning \cite{balestriero2023cookbook} on a huge amount of unlabeled data \cite{Zhu-ICCV-2015-Aligning, Liu-CoRR-2019-RoBERTa, Zellers-NeurIPS-2019-Defending, Gokaslan2019OpenWeb}, which cover a wide range of genres, e.g., encyclopedias, news, books, social medias, etc. Many studies have demonstrated that LLMs are able to acquire broad knowledge of many types and subjects \cite{LLMSurvey, Paperno-ACL-2016-LAMBADA, Hoffmann-arxiv-2022-Training, Touvron-arxiv-2023-LLaMA, Rae-arxiv-2021-Scaling, Raffel-JMLR-2020-Exploring, Du-ICML-2022-GLaM}.

The paradigms that elicit and apply the acquired knowledge in LLMs onto downstream tasks have shifted from fine-tuning to instruction-tuning.  Early LLMs usually adopt fine-tuning, which, however, suffers from lack of cross-task generalization as the fine-tuned LLMs are often task-specific and not being parameter-efficient as all pre-trained LLM parameters are usually required to be updated on downstream tasks. As LLMs reach the scale of billions of parameters, a more efficient alternative to elicit knowledge, in-context Learning (ICL) \cite{Brown-NeurIPS-2020-Language, Michael-ICLR-2022-An, Dong-arxiv-2023-A} has emerged, which uses only a few demonstration examples concatenated in a prompt. In order to enhance the cross-task generalization of LLMs to a variety of downstream tasks, instruction-tuning \cite{Wei-ICLR-2022-Finetuned, Bach-ACL-2022-PromptSource, Wang-EMNLP-2022-Super}, which is performed via multi-task learning \cite{Chung-arxiv-2022-Scaling, Liu-ACL-2019-Multi} has been proposed. In instruction-tuning, the instructions for different tasks are different, but in a unified form. Supervised Fine-tuning (SFT) \cite{Ouyang-arxiv-2022-Training} and Reinforcement Learning from Human Feedback (RLHF) \cite{Christiano-NeurIPS-2017-Deep, Stiennon-arxiv-2020-learning, Ouyang-arxiv-2022-Training} are successful methods of instruction-tuning, which not only achieve generalization to unseen instructions but also align LLMs with human values and intents \cite{Sanh-ICLR-2022-Multitask, Wei-ICLR-2022-Finetuned, Chung-arxiv-2022-Scaling}.


As the capability of knowledge acquisition and application in LLMs is constantly and rapidly evolving, a natural question which arises, is how we can assess such knowledge. Traditional single-task evaluation benchmarks \cite{Rajpurkar-EMNLP-2016-SQuAD, Khot-AAAI-2020-QASC} are no longer adequate for evaluating them. Multi-task benchmarks like GLUE \cite{Wang-EMNLP-2018-GLUE}, SuperGLUE \cite{Wang-NIPS-2019-SuperGLUE} and BIG-bench \cite{Srivastava-arxiv-2022-Beyond} aggregate multiple NLP tasks to evaluate LLMs, which, however, are not sufficient either to assess knowledge acquired by LLMs. To address this issue, \citet{Hendrycks-ICLR-2021-Measuring} propose MMLU, a widely used benchmark to test the knowledge acquisition and application capability of LLMs, which uses test questions across multiple subjects that humans lean to assess LLMs in zero- and few-shot settings. As MMLU is an English benchmark, it cannot be directly used for measuring LLMs trained with data in other languages. Even if it is translated into other languages, like the way used in evaluating GPT-4 \cite{OpenAI-OpenAI-2023-GPT-4}, there are still gaps in knowledge across different languages as they usually have different education systems and knowledge structures.

\begin{table}
\centering
\resizebox{0.48\textwidth}{!}
{
\begin{tabular}{lccc}
\hline
\textbf{Benchmark} & \textbf{Language}& \textbf{\# Tasks} & \textbf{\# Questions}\\
\hline
    \text{MMLU \cite{Hendrycks-ICLR-2021-Measuring}} & \text{En} & \text{57} & \text{15,908}\\
    \text{AGIEval \cite{zhong2023agieval}} & \text{En \& Zh} & \text{20} & \text{8,062}\\
    \text{MMCU \cite{zeng2023measuring}} & \text{Zh} & \text{51} & \text{11,900}\\
    \text{M3KE} & \text{Zh} & \text{71} & \text{20,477}\\
\hline
\end{tabular}
}
\caption{The comparison between M3KE and other related benchmarks.}
\label{compared}
\end{table}

Similar to LLMs in English, LLMs dedicated in Chinese have also achieved rapid advances recently \cite{du-etal-2022-glm, Zeng-arxiv-2021-PanGualpha, Zhang-arXiv-2021-CPM-2, Sun-arXiv-2021-ERNIE3.0, Zeng-arxiv-2022-GLM, Ren-arXiv-2023-PanGusigma, Wu-arxiv-2021-Yuan, Wang-arxiv-2021-ERNIE, chen2023phoenix}. However, a massive knowledge evaluation benchmark that measures Chinese LLMs in line with Chinese education system is a desideratum.  To bridge this gap, we propose M3KE, a Massive Multi-Level Multi-Subject Knowledge Evaluation benchmark, which is designed to measure the knowledge acquired by Chinese LLMs by testing their multitask accuracy in zero- and few-shot settings. M3KE contains 20,477 questions collected from 71 tasks. In particular, unlike recent benchmarks MMCU \cite{zeng2023measuring} and AGIEval \cite{zhong2023agieval}, M3KE covers all major levels of Chinese education system, ranging from primary school to college, as well as a wide variety of subjects, including humanities, history, politics, law, education, psychology, science, technology, art and religion. All questions are multiple-choice questions with four options, hence ensuring a standardized and unified assessment process. Table~\ref{compared} shows the comparison between M3KE and other related benchmarks.

With M3KE, we have tested recently released Chinese LLMs , to track the progress of Chinese LLMs in knowledge acquisition and application. The evaluated models are either pre-trained on massive data or pre-trained + fine-tuned with SFT or RLHF. The model sizes vary from 335M to 130B parameters.

With extensive experiments, we observe that most evaluated Chinese LLMs have near random-chance accuracy, even for primary school tasks. The best performance is achieved by an SFT model built on the open-source BLOOM \cite{Scao-arxiv-2022-BLOOM}, which is 14.8 points lower than the accuracy of GPT-3.5-turbo.

Our main contributions are summarized as follows.
\begin{itemize}
\item We propose M3KE, a knowledge evaluation benchmark for Chinese LLMs, which to date covers the largest number of tasks in line with Chinese education system. 
\item  We have tested a wide range of open-source Chinese LLMs, with model sizes varying from 335M to 130B, against GPT-3.5-turbo.
\item We have analyzed the performance of each model on different subject clusters and education levels in both zero- and five-shot settings.
\end{itemize}

\section{Related Work}
\paragraph{Chinese Large Language Models.}

Recent years have witnessed a rapid development of Chinese LLMs, following the efforts of their English counterparts, e.g., GPT-3 \cite{Brown-NeurIPS-2020-Language}, Gopher \cite{Rae-arxiv-2021-Scaling}, LLaMA \cite{Touvron-arxiv-2023-LLaMA}. Chinese LLMs, such as Pangu-$\alpha$ with 200B parameters \cite{Zeng-arxiv-2021-PanGualpha}, Yuan 1.0 with 245B parameters \cite{Wu-arxiv-2021-Yuan}, ERNIE 3.0 Titan with 260B parameters \cite{Sun-arXiv-2021-ERNIE3.0}, have been trained on Chinese textual data that contain tokens ranging from 180B to 329B. These models are developed in industry, which are usually not open-source. With the success of open-source LLMs \cite{alpaca, peng2023instruction} based on LLaMA, Chinese versions, such as ChatGLM-6B\footnote{https://github.com/THUDM/ChatGLM-6B}, MOSS\footnote{https://github.com/OpenLMLab/MOSS}, Phoenix \cite{chen2023phoenix}, have emerged very recently. These models usually contain less than 20 billion parameters and are supervised fine-tuned on instructions that are either distilled from models of GPT-3.5 or learned in a self-instructing manner \cite{Wang-arXiv-2022-Self}.


\paragraph{Benchmarks.}

The capability of eliciting and applying knowledge acquired during training is an important indicator for measuring LLMs. However, existing evaluation benchmarks \cite{Wang-EMNLP-2018-GLUE, Wang-NIPS-2019-SuperGLUE, Srivastava-arxiv-2022-Beyond, Xu-COLING-2020-CLUE} are normally designed to evaluate LLMs on various NLP tasks, not tailored for knowledge acquisition and application assessment. To comprehensively measure knowledge in LLMs, MMLU \cite{Hendrycks-ICLR-2021-Measuring} is proposed, which collects multiple-choice questions from 57 tasks that humans learn. As a different education system is used, on the one side, knowledge in Chinese LLMs may not exhibit in the translated-into-Chinese version of MMLU, e.g., Chinese Medicine, Chinese Legal System. On the other side, knowledge to be assessed in MMLU may be absent in Chinese textual data used to train Chinese LLMs. 

Our work is related to 3 datasets that have been developed concurrently with M3KE. MMCU \cite{zeng2023measuring} is a Chinese benchmark that assesses knowledge in four domains: medicine, education, law, and psychology. AGIEval \cite{zhong2023agieval} is a bilingual benchmark that measures the capability of LLMs on tasks of the Chinese college entrance exam and American college admission test, for high-school graduates. DomMa \cite{gu2023domain} is another Chinese benchmark that focuses on domain-specific knowledge. In contrast to these benchmarks, M3KE is a comprehensive Chinese benchmark that spans major stages of Chinese education system, from primary school to college with a broader range of subject categories, such as art, religion, traditional Chinese medicine, and classical literature.

\begin{figure}
\centering
    \includegraphics[width=0.5\textwidth]{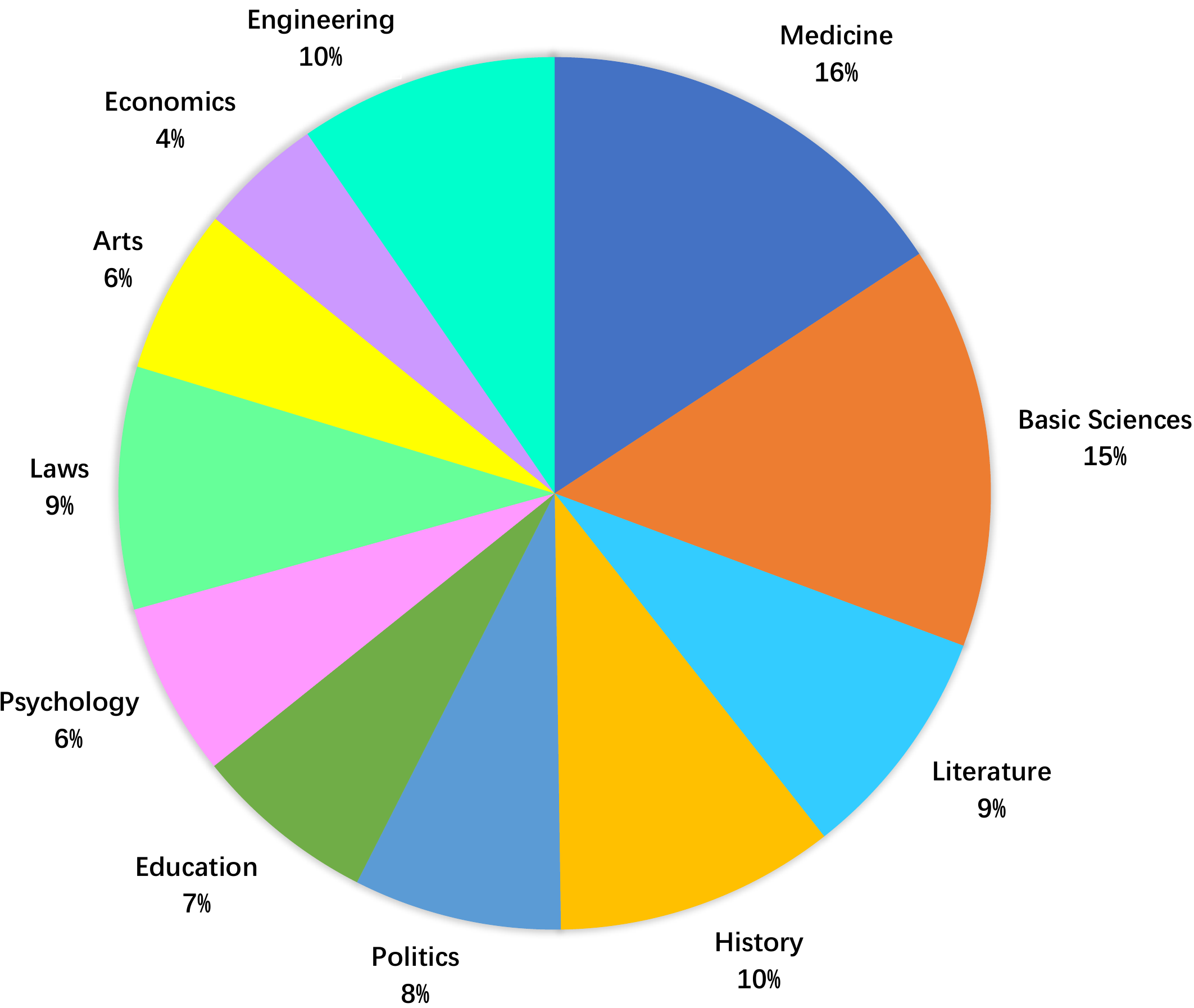}
    \caption{The distribution of tasks in M3KE.} \label{contribution}
\end{figure}

\section{M3KE}
M3KE covers major Chinese education levels, including primary school, middle school, high school, college and professional exams, as well as multiple tasks as shown in Figure~\ref{contribution} while the detailed subjects are listed in Appendix ~\ref{sec:subjects}. We collect and organize multiple-choice questions from public websites. To ensure the quality and comprehensiveness of the questions, entrance exam questions are selected as much as possible. For the primary school, middle school and high school education level, we choose the subjects according to the corresponding entrance exams for Chinese students. For the college level, we select subjects according to the national entrance exam for master's degree in China. In addition to subjects under the major Chinese education system, we also collect comprehensive tasks to expand the knowledge coverage in M3KE, including computer grade exam, ancient Chinese language, novels and Chinese national civil service exam which covers commonsense knowledge, arts, religion, etc.

In total, we have 71 tasks and 20,477 questions. We divide each task into a test set and a few-shot set, where the few-shot set includes 5 questions for each task for the few-shot evaluation setting. The test set includes 20,122 questions, and each task contains at least 100 questions. Instances of M3KE are listed in Table~\ref{Other}.

\begin{table*}
\centering
\small
\begin{tabularx}{\textwidth}{llXX}
\toprule
\multirow{5}{*}{Arts \& Humanities } &
&\textbf{下面关于拉斯科洞穴壁画说法错误的是?}&\textbf{Which statement about the Lascaux cave murals is incorrect?}  \\
\cmidrule{2-4}
 & A & 这个壁画是在法国发现的 & This fresco was found in France\\
 & B & 发现的动物形象有100多个 & There are more than 100 animal images found\\
 & C & 发现的时间为1940年 & The discovery was made in 1940\\
 & \textbf{D}& \textbf{壁画颜色以黑色为主} & \textbf{Mural color is mainly black}\\
\midrule
\multirow{5}{*}{Social Sciences} & &\textbf{甲欲杀乙,将毒药投入乙的饭食中. 乙服食后,甲后悔,赶紧说明情况,并将乙送往医院抢救.医院在抢救过程中检查发现,甲所投放的"毒药"根本没有毒性,乙安然无恙.甲的行为属于?}&\textbf{A wants to kill B, and puts poison into B's food. After B consumed it, A regretted it and rushed to explain the situation and sent B to the hospital for rescue. The hospital found that the poison was not toxic at all and B was unharmed. A's behavior belongs to?}  \\
\cmidrule{2-4}
& A & 不构成犯罪 & Not a crime\\
& B & 犯罪未遂 & Attempted crime\\
& \textbf{C}&\textbf{犯罪中止} & \textbf{Crime suspension}\\
& D & 犯罪既遂 & Crime reached\\
\midrule
\multirow{5}{*}{Natural Sciences} & &\textbf{使用普鲁卡因麻醉神经纤维,影响了神经纤维传导兴奋的哪一项特征?}&\textbf{Which characteristic of nerve fiber conduction excitation is affected by the use of procaine anesthesia?}  \\
\cmidrule{2-4}
& \textbf{A}&\textbf{生理完整性} & \textbf{Physiological integrity}\\
& B & 绝缘性 & Insulation \\
& C & 双向传导性 & Bidirectional conduction\\
& D & 相对不疲劳性 & Relative non-fatigability\\
\midrule
\multirow{5}{*}{Other} &  &\textbf{以前有几项研究表明，食用巧克力会增加食用者患心脏病的可能性。而一项最新的、更为可靠的研究得出的结论是：食用巧克力与心脏病发病率无关。估计这项研究成果公布以后，巧克力的消费量将会大大增加。上述推论基于以下哪项假设?}&\textbf{Several studies have previously suggested that consuming chocolate increases the likelihood of developing heart disease. However, a recent and more reliable study concluded that there is no association between chocolate consumption and incidence of heart disease. It is estimated that the consumption of chocolate will significantly increase after the publication of this research. The above inference is based on the assumption that the reliability of the previous studies was lower than that of the latest study.}  \\
\cmidrule{2-4}
& A & 尽管有些人知道食用巧克力会增加患心脏病的可能性，却照样大吃特吃 & Although some people are aware that consuming chocolate increases the likelihood of developing heart disease, they still indulge in it.\\
& B & 人们从来也不相信进食巧克力会更容易患心脏病的说法 & People have never believed the claim that eating chocolate makes it more likely to develop heart disease.\\
& C &现在许多人吃巧克力是因为他们没有听过巧克力会导致心脏病的说法 & Nowadays, many people eat chocolate because they have not heard of the claim that chocolate can lead to heart disease.\\
& \textbf{D}&\textbf{现在许多人不吃巧克力完全是因为他们相信巧克力会诱发心脏病} & \textbf{Nowadays, many people abstain from eating chocolate solely because they believe that chocolate can trigger heart disease.}\\
\bottomrule
\end{tabularx}
\caption{\label{Other}
Examples from M3KE. Bolded items represent correct answers. Examples from top to bottom are from Fine Arts, Criminal Jurisprudence, Animal Physiology and Chinese Civil Service Examination task, respectively. }
\end{table*}

\begin{table*}
	\centering
	\begin{tabular}{lcccc}
		\hline &\textbf{Arts \& Humanities}& \textbf{Social Sciences} & \textbf{Natural Sciences} &\textbf{Other} \\ \hline
		Tasks & 12 &21 & 31&7 \\
        Q Numbers & 3,612 &6,222&8,162&2,126 \\
        Avg.Q Numbers &301&296&263&303 \\
        Max.Q Numbers & 352 & 374&347&425 \\
        Min.Q Numbers & 190 & 190&100&129 \\
		Avg.Q Tokens  & 30.33 &38.75&38.54 & 33.21 \\
		Avg.C Tokens & 53.92&30.99 & 44.57& 52.53\\
		\hline
	\end{tabular}
	\caption{\label{stacft} Overall statistics of M3KE. Q: question. C: answer choices }
\end{table*}

\subsection{Arts \& Humanities}
Arts \& Humanities comprise a range of disciplines that cover Chinese, literature, arts and history. These disciplines focus on the analysis and interpretation of literary and cultural artifacts, rather than on practical applications. For instance, the Chinese in primary school aims to evaluate the students' proficiency in language use and literary appreciation for ages 7 to 13, such as the usage of synonyms and antonyms. The historical studies cover both Chinese and world history from ancient to modern times. M3KE also incorporates artistic subjects, such as dance, fine arts, music and film, because we believe that art is an essential aspect of human culture and should be relevant to LLMs as well.

\subsection{Social Sciences}
Social sciences differ from Arts \& Humanities in that they emphasize practical aspects of humanistic studies, such as law, politics, education and psychology. These subjects are mainly taught at the college level. Although ideological and political courses are also part of the Chinese middle school and high school curriculum, they primarily involve moral education. Social sciences also encompass economic and management studies, which largely consist of questions from the joint exams for graduate students majoring in these fields in China. These studies include microeconomics, macroeconomics, management and logic at the undergraduate level. 

\subsection{Natural Sciences}
Natural sciences encompass engineering, science, medicine and fundamental disciplines such as math, physics, chemistry, biology and so on. These subjects often require a high degree of computation, analysis and logical reasoning skills. The same subject may assess different types of knowledge at different levels according to the Chinese education system. For instance, primary school math mainly tests the basic arithmetic operations, while high school math covers more advanced mathematical concepts, such as sequences, derivatives and geometry.

\subsection{Other}
Other types of tasks include religion, Chinese civil service exam, and specialized tasks, like ancient Chinese language and novel reasoning task. These tasks require knowledge that is not limited to a single level or subject as described above. The Chinese civil service exam involves knowledge in commonsense, humanities, logic and other domains, which we can consider as an assessment of the comprehensive knowledge for LLMs. Similarly, in the novel task, these questions involve a lot of information from many classical novels.

\begin{table*}
\centering
\begin{tabular}{lccccc}
\hline
\text{Models}&\text{Arts \& Humanities}&\text{Social Sciences}&\text{Natural Sciences}&\text{Other}&\text{Average} \\
\hline
\text{GLM-335M}&\text{0.070} &\text{0.046} &\text{0.084}&\text{0.044} &\text{0.062}\\
\text{BLOOM-7.1B}&\text{0.163} & \text{0.159}&\text{0.161} &\text{0.158}&\text{0.161}\\
\text{GLM-10B}&\text{0.180} &\text{0.229} &\text{0.219}&\text{0.150} &\text{0.197}\\
\text{GLM-130B}&\text{0.326} &\text{0.352} &\text{0.274}&\text{0.359} &\text{0.328}\\
\hline
\text{ChatGLM-6B}&\text{0.246} &\text{0.267} &\text{0.168}&\text{0.263} &\text{0.236}\\
\text{MOSS-SFT-16B}&\text{0.260} &\text{0.263} &\text{0.207}&\text{0.275} &\text{0.251}\\
\text{BELLE-7B-0.2M}&\text{0.247}&\text{0.296} &\text{0.260}&\text{0.260} &\text{0.266}\\
\text{BELLE-7B-2M}&\text{0.328} &\text{0.367} &\text{0.282}&\text{0.355} &\text{0.333}\\
\text{GPT-3.5-turbo}&\text{0.460} &\text{0.538} &\text{0.444}&\text{0.481} &\text{0.481}\\
\hline
\end{tabular}
\caption{\label{zero-shot}
Average zero-shot accuracy for each model on the four subject clusters.
}
\end{table*}

\subsection{Overall Statistics}
Table~\ref{stacft} shows the overall statistics of M3KE. The numbers of tasks in the four subject clusters described above are 12, 21, 31 and 7, respectively, while the numbers of questions in the four subject clusters are 3,612, 6,222, 8,162 and 2,126, respectively. The maximum number of questions is 425 while the minimum number is 100. Questions in social and natural sciences are usually longer than those in arts \& humanities and other while their answer choices are shorter.  

\begin{table*}
\centering
\begin{tabular}{lccccc}
\hline
\text{Models}&\text{Arts \& Humanities}&\text{Social Sciences}&\text{Natural Sciences}&\text{Other}&\text{Average} \\
\hline
\text{GLM-335M}&\text{0.220}&\text{0.247}&\text{0.193}&\text{0.126}&\text{0.196}\\
\text{BLOOM-7.1B}&\text{0.247} & \text{0.260}&\text{0.235} &\text{0.246}&\text{0.247}\\
\text{GLM-10B}& \text{0.294}&\text{0.304}&\text{0.232}&\text{0.211}&\text{0.260}\\
\text{GLM-130B}&\text{0.297}&\text{0.329}&\text{0.246}&\text{0.228}&\text{0.275}\\
\hline
\text{ChatGLM-6B}&\text{0.188}&\text{0.175}&\text{0.121}&\text{0.198}&\text{0.171}\\
\text{MOSS-SFT-16B}&\text{0.266}&\text{0.264}&\text{0.258}&\text{0.284}&\text{0.268}\\
\text{BELLE-7B-0.2M} & \text{0.292}&\text{0.327}&\text{0.273}&\text{0.307}&\text{0.299}\\
\text{BELLE-7B-2M} & \text{0.287} &\text{0.309}&\text{0.284} &\text{0.313}&\text{0.298}\\
\text{GPT-3.5-turbo}&\text{0.453} &\text{0.540} &\text{0.464}&\text{0.476} &\text{0.483}\\
\hline
\end{tabular}
\caption{\label{5 shot}
Average five-shot accuracy for each model on the four subject clusters.
}
\end{table*}

\section{Experiments}

We assessed state-of-the-art large language models recently developed for Chinese on M3KE, attempting to understand and track the progress of Chinese LLMs in learning and applying knowledge from massive data. 

\subsection{Assessed Models}
The assessed Chinese LLMs can be divided into two categories: models being only pre-trained and models that are instruction-tuned with SFT/RLHF. For the former, we selected GLM-335M \cite{du-etal-2022-glm}, GLM-10B \cite{du-etal-2022-glm}, GLM-130B \cite{Zeng-arxiv-2022-GLM} and BLOOM-7.1B \cite{Scao-arxiv-2022-BLOOM}. For the latter, we included ChatGLM-6B\footnote{https://github.com/THUDM/ChatGLM-6B}, MOSS-SFT-16B\footnote{https://huggingface.co/fnlp/moss-moon-003-sft}, BELLE-7B \cite{BELLE}, where BELLE-7B is the SFT version based on BLOOMZ-7.1B-MT \cite{Muennighoff-2022-arxiv-Crosslingual}. We used the two variants of BELLE fine-tuned on 200K and 2M instructions, namely BELLE-7B-0.2M\footnote{https://huggingface.co/BelleGroup/BELLE-7B-0.2M} and BELLE-7B-2M\footnote{https://huggingface.co/BelleGroup/BELLE-7B-2M}. We also evaluated GPT-3.5-turbo\footnote{https://openai.com/product} from OpenAI as a reference.

\begin{table*}
\centering
\begin{tabular}{lcccccc}
\hline
\text{Models}&\text{Primary School}&\text{Middle School}&\text{High School}&\text{College}&\text{Other}&\text{Average} \\
\hline
\text{GLM-335M}&\text{0.075} &\text{0.099} &\text{0.099}&\text{0.054} &\text{0.046}&\text{0.075}\\
\text{BLOOM-7.1B}&\text{0.173} & \text{0.142}&\text{0.173} &\text{0.160}&\text{0.164}&\text{0.163}\\
\text{GLM-10B}&\text{0.190} &\text{0.199} &\text{0.197}&\text{0.213} &\text{0.152}&\text{0.190}\\
\text{GLM-130B}&\text{0.243} &\text{0.303} &\text{0.229}&\text{0.324} &\text{0.359}&\text{0.292}\\
\hline
\text{ChatGLM-6B}&\text{0.180} &\text{0.243} &\text{0.191}&\text{0.213} &\text{0.250}&\text{0.216}\\
\text{MOSS-SFT-16B}&\text{0.224} &\text{0.223} &\text{0.213}&\text{0.242} &\text{0.260}&\text{0.232}\\
\text{BELLE-7B-0.2M}&\text{0.233}&\text{0.269} &\text{0.259}&\text{0.268} &\text{0.263} &\text{0.258}\\
\text{BELLE-7B-2M}&\text{0.248} &\text{0.313} &\text{0.263}&\text{0.332} &\text{0.349}&\text{0.301}\\
\text{GPT-3.5-turbo}&\text{0.328} &\text{0.403} &\text{0.395}&\text{0.509} &\text{0.484}&\text{0.435}\\
\hline
\end{tabular}
\caption{\label{zero-shot edu}
Average zero-shot accuracy for each model on five major education levels.
}
\end{table*}

\begin{table*}
\centering
\begin{tabular}{lcccccc}
\hline
\text{Models}&\text{Primary School}&\text{Middle School}&\text{High School}&\text{College}&\text{Other}&\text{Average} \\
\hline
\text{GLM-335M}&\text{0.206}&\text{0.229}&\text{0.232}&\text{0.223}&\text{0.114}&\text{0.201}\\
\text{BLOOM-7.1B}&\text{0.262} & \text{0.222}&\text{0.245} &\text{0.249}&\text{0.246}&\text{0.245}\\
\text{GLM-10B}& \text{0.229}&\text{0.263}&\text{0.270}&\text{0.278}&\text{0.197}&\text{0.248}\\
\text{GLM-130B}&\text{0.268}&\text{0.293}&\text{0.272}&\text{0.294}&\text{0.208}&\text{0.267}\\
\hline
\text{ChatGLM-6B}&\text{0.089}&\text{0.150}&\text{0.137}&\text{0.155}&\text{0.196}&\text{0.146}\\
\text{MOSS-SFT-16B}&\text{0.272}&\text{0.223}&\text{0.263}&\text{0.266}&\text{0.281}&\text{0.261}\\
\text{BELLE-7B-0.2M} & \text{0.260}&\text{0.256}&\text{0.273}&\text{0.298}&\text{0.310}&\text{0.280}\\
\text{BELLE-7B-2M} & \text{0.258} &\text{0.264}&\text{0.268} &\text{0.306}&\text{0.299}&\text{0.279}\\
\text{GPT-3.5-turbo}&\text{0.308} &\text{0.565} &\text{0.373}&\text{0.517} &\text{0.475}&\text{0.448}\\
\hline
\end{tabular}
\caption{\label{5 shot edu} 
Average five-shot accuracy for each model on five major education levels.
}
\end{table*}

\subsection{Prompts}
All models were tested using the $n$-shot setting with a unified prompt, where $n$ is an integer from 0 to 5. For the zero-shot setting  (i.e., $n=0$), the unified prompt provided to all models is ``Please choose the correct option from `A', `B', `C', `D' based on the following question''. For few-shot setting (i.e., $n > 0$), the unified prompt is ``Please choose the correct option from `A', `B', `C', `D' based on the following examples and question''. The input to all LLMs consists of the prompt, question, answer choices and suffix, which is ``the correct option is: ''. Even we tell models to only output the correct answer choice indicator (i.e., $\in \{A, B, C, D\}$) in the prompt, not all models can follow this instruction. Sometimes they output both answer choice and rationale to the answer choice (the order of these two types of outputs are random). We hence keep only the output answer choice indicator as the final answer to calculate accuracy. 

\subsection{Results}
We compared the zero-shot accuracy of each model in Table~\ref{zero-shot} in terms of subject clusters. For the pre-trained models, there is a clear positive correlation between accuracy and model size, where the model with 130B parameters significantly outperforms the models with 335M/7B/10B parameters, even though they have different backbones. The accuracy of GPT-3.5-turbo is significantly higher than those of the evaluated Chinese LLMs, which currently provides an upper bound for open-source Chinese LLMs. All pretrained LLMs with $\leq 10B$ parameters achieve an accuracy lower than random-chance accuracy (i.e., 25$\%$), indicating that knowledge acquired by these models is not adequate for M3KE. In addition, we observe that the number of instructions used for SFT is an important factor, as the BELLE model fine-tuned with 2M instructions is significantly better than that with 0.2M instructions. The zero-shot performance of GPT-3.5-turbo is much higher than the compared open-sourced Chinese LLMs, but still lower than 50$\%$ accuracy, suggesting that M3KE is a very challenging benchmark. 

We further compared the accuracy of different models under the 5-shot setting. Results are shown in Table~\ref{5 shot}. For pre-trained models, ICL in the few-shot setting significantly improves the performance and the smaller the pretrained model is, the larger the achieved improvement is. The exception is GLM-130B, which performs significantly worse under the 5-shot setting than the zero-shot setting. We conjecture that GLM-130B already has the ability to understand questions without examples because it uses instances in the instruction format as part of the pre-training corpus \cite{Zeng-arxiv-2022-GLM}, and demonstrations may bring interference to the final prediction of the model. The 5-shot results of the SFT models are mixed in comparison to those in the zero-shot setting. We find that for ChatGLM-6B and BELLE-7B-2M, 5-shot is worse than zero-shot setting, similar to the results observed on GLM-130B. In contrast, 5-shot has a positive impact on MOSS-SFT-16B and BELLE-7B-0.2M. As these models are different from each other in terms of model size, training data, instruction data, etc., we leave the in-depth analysis on the mixed results to our future work. 

We finally provide the results of each model on different education levels in Table~\ref{zero-shot edu} for the zero-shot setting and Table~\ref{5 shot edu} for the few-shot setting. Interestingly, we observe that LLMs do not reach higher performance at lower education levels than higher education levels, even for GPT-3.5-turbo. This suggests that tasks from lower education levels remain challenging for these state-of-the-art Chinese LLMs.



\section{Conclusion}
We have presented a new benchmark M3KE, to assess the capability of Chinese LLMs in learning and applying knowledge in multiple subjects at multiple levels of Chinese education system. M3KE contains 71 tasks and 20,447 questions. We find that all evaluated state-of-the-art open-source Chinese LLMs significantly lag behind GPT-3.5. We hope that this benchmark can be used to track and promote further progress in Chinese LLMs.

\end{CJK}
\newpage
\bibliography{anthology,custom}

\begin{thebibliography}{57}
\expandafter\ifx\csname natexlab\endcsname\relax\def\natexlab#1{#1}\fi

\bibitem[{Altman(2023)}]{OpenAI-blog-2023-Planning}
Sam Altman. 2023.
\newblock Planning for agi and beyond.
\newblock \emph{OpenAI Blog}.

\bibitem[{Bach et~al.(2022)Bach, Sanh, Yong, Webson, Raffel, Nayak, Sharma,
  Kim, Bari, F{\'{e}}vry, Alyafeai, Dey, Santilli, Sun, Ben{-}David, Xu,
  Chhablani, Wang, Fries, AlShaibani, Sharma, Thakker, Almubarak, Tang, Radev,
  Jiang, and Rush}]{Bach-ACL-2022-PromptSource}
Stephen~H. Bach, Victor Sanh, Zheng~Xin Yong, Albert Webson, Colin Raffel,
  Nihal~V. Nayak, Abheesht Sharma, Taewoon Kim, M.~Saiful Bari, Thibault
  F{\'{e}}vry, Zaid Alyafeai, Manan Dey, Andrea Santilli, Zhiqing Sun, Srulik
  Ben{-}David, Canwen Xu, Gunjan Chhablani, Han Wang, Jason~Alan Fries,
  Maged~Saeed AlShaibani, Shanya Sharma, Urmish Thakker, Khalid Almubarak,
  Xiangru Tang, Dragomir~R. Radev, Mike~Tian{-}Jian Jiang, and Alexander~M.
  Rush. 2022.
\newblock Promptsource: An integrated development environment and repository
  for natural language prompts.
\newblock In \emph{{ACL} (demo)}, pages 93--104. Association for Computational
  Linguistics.

\bibitem[{Balestriero et~al.(2023)Balestriero, Ibrahim, Sobal, Morcos, Shekhar,
  Goldstein, Bordes, Bardes, Mialon, Tian et~al.}]{balestriero2023cookbook}
Randall Balestriero, Mark Ibrahim, Vlad Sobal, Ari Morcos, Shashank Shekhar,
  Tom Goldstein, Florian Bordes, Adrien Bardes, Gregoire Mialon, Yuandong Tian,
  et~al. 2023.
\newblock A cookbook of self-supervised learning.
\newblock \emph{arXiv preprint arXiv:2304.12210}.

\bibitem[{Brown et~al.(2020)Brown, Mann, Ryder, Subbiah, Kaplan, Dhariwal,
  Neelakantan, Shyam, Sastry, Askell, Agarwal, Herbert{-}Voss, Krueger,
  Henighan, Child, Ramesh, Ziegler, Wu, Winter, Hesse, Chen, Sigler, Litwin,
  Gray, Chess, Clark, Berner, McCandlish, Radford, Sutskever, and
  Amodei}]{Brown-NeurIPS-2020-Language}
Tom~B. Brown, Benjamin Mann, Nick Ryder, Melanie Subbiah, Jared Kaplan,
  Prafulla Dhariwal, Arvind Neelakantan, Pranav Shyam, Girish Sastry, Amanda
  Askell, Sandhini Agarwal, Ariel Herbert{-}Voss, Gretchen Krueger, Tom
  Henighan, Rewon Child, Aditya Ramesh, Daniel~M. Ziegler, Jeffrey Wu, Clemens
  Winter, Christopher Hesse, Mark Chen, Eric Sigler, Mateusz Litwin, Scott
  Gray, Benjamin Chess, Jack Clark, Christopher Berner, Sam McCandlish, Alec
  Radford, Ilya Sutskever, and Dario Amodei. 2020.
\newblock Language models are few-shot learners.
\newblock In \emph{Advances in Neural Information Processing Systems 33: Annual
  Conference on Neural Information Processing Systems 2020, NeurIPS 2020,
  December 6-12, 2020, virtual}.

\bibitem[{Bubeck et~al.(2023)Bubeck, Chandrasekaran, Eldan, Gehrke, Horvitz,
  Kamar, Lee, Lee, Li, Lundberg, Nori, Palangi, Ribeiro, and
  Zhang}]{Bubeck-arxiv-2023-Sparks}
S'ebastien Bubeck, Varun Chandrasekaran, Ronen Eldan, Johannes Gehrke, Eric
  Horvitz, Ece Kamar, Peter Lee, Yin~Tat Lee, Yuanzhi Li, Scott Lundberg,
  Harsha Nori, Hamid Palangi, Marco~Tulio Ribeiro, and Yi~Zhang. 2023.
\newblock Sparks of artificial general intelligence: Early experiments with
  gpt-4.
\newblock volume abs/2303.12712.

\bibitem[{Cao et~al.(2023)Cao, Li, Liu, Yan, Dai, Yu, and
  Sun}]{Cao-arxiv-2023-comprehensive}
Yihan Cao, Siyu Li, Yixin Liu, Zhiling Yan, Yutong Dai, Philip~S Yu, and Lichao
  Sun. 2023.
\newblock A comprehensive survey of ai-generated content (aigc): A history of
  generative ai from gan to chatgpt.
\newblock \emph{arXiv preprint arXiv:2303.04226}.

\bibitem[{Chen et~al.(2023)Chen, Jiang, Chen, Wang, Yu, Chen, Zhang, Liang,
  Zhang, Zhang et~al.}]{chen2023phoenix}
Zhihong Chen, Feng Jiang, Junying Chen, Tiannan Wang, Fei Yu, Guiming Chen,
  Hongbo Zhang, Juhao Liang, Chen Zhang, Zhiyi Zhang, et~al. 2023.
\newblock Phoenix: Democratizing chatgpt across languages.
\newblock \emph{arXiv preprint arXiv:2304.10453}.

\bibitem[{Christiano et~al.(2017)Christiano, Leike, Brown, Martic, Legg, and
  Amodei}]{Christiano-NeurIPS-2017-Deep}
Paul~F. Christiano, Jan Leike, Tom~B. Brown, Miljan Martic, Shane Legg, and
  Dario Amodei. 2017.
\newblock Deep reinforcement learning from human preferences.
\newblock In \emph{Advances in Neural Information Processing Systems 30: Annual
  Conference on Neural Information Processing Systems 2017, December 4-9, 2017,
  Long Beach, CA, {USA}}, pages 4299--4307.

\bibitem[{Chung et~al.(2022)Chung, Hou, Longpre, Zoph, Tay, Fedus, Li, Wang,
  Dehghani, Brahma, Webson, Gu, Dai, Suzgun, Chen, Chowdhery, Narang, Mishra,
  Yu, Zhao, Huang, Dai, Yu, Petrov, Chi, Dean, Devlin, Roberts, Zhou, Le, and
  Wei}]{Chung-arxiv-2022-Scaling}
Hyung~Won Chung, Le~Hou, Shayne Longpre, Barret Zoph, Yi~Tay, William Fedus,
  Eric Li, Xuezhi Wang, Mostafa Dehghani, Siddhartha Brahma, Albert Webson,
  Shixiang~Shane Gu, Zhuyun Dai, Mirac Suzgun, Xinyun Chen, Aakanksha
  Chowdhery, Sharan Narang, Gaurav Mishra, Adams Yu, Vincent~Y. Zhao, Yanping
  Huang, Andrew~M. Dai, Hongkun Yu, Slav Petrov, Ed~H. Chi, Jeff Dean, Jacob
  Devlin, Adam Roberts, Denny Zhou, Quoc~V. Le, and Jason Wei. 2022.
\newblock Scaling instruction-finetuned language models.
\newblock \emph{CoRR}, abs/2210.11416.

\bibitem[{Dong et~al.(2023)Dong, Li, Dai, Zheng, Wu, Chang, Sun, Xu, Li, and
  Sui}]{Dong-arxiv-2023-A}
Qingxiu Dong, Lei Li, Damai Dai, Ce~Zheng, Zhiyong Wu, Baobao Chang, Xu~Sun,
  Jingjing Xu, Lei Li, and Zhifang Sui. 2023.
\newblock A survey for in-context learning.
\newblock \emph{CoRR}, abs/2301.00234.

\bibitem[{Du et~al.(2022{\natexlab{a}})Du, Huang, Dai, Tong, Lepikhin, Xu,
  Krikun, Zhou, Yu, Firat, Zoph, Fedus, Bosma, Zhou, Wang, Wang, Webster,
  Pellat, Robinson, Meier{-}Hellstern, Duke, Dixon, Zhang, Le, Wu, Chen, and
  Cui}]{Du-ICML-2022-GLaM}
Nan Du, Yanping Huang, Andrew~M. Dai, Simon Tong, Dmitry Lepikhin, Yuanzhong
  Xu, Maxim Krikun, Yanqi Zhou, Adams~Wei Yu, Orhan Firat, Barret Zoph, Liam
  Fedus, Maarten~P. Bosma, Zongwei Zhou, Tao Wang, Yu~Emma Wang, Kellie
  Webster, Marie Pellat, Kevin Robinson, Kathleen~S. Meier{-}Hellstern, Toju
  Duke, Lucas Dixon, Kun Zhang, Quoc~V. Le, Yonghui Wu, Zhifeng Chen, and
  Claire Cui. 2022{\natexlab{a}}.
\newblock Glam: Efficient scaling of language models with mixture-of-experts.
\newblock In \emph{International Conference on Machine Learning, {ICML} 2022,
  17-23 July 2022, Baltimore, Maryland, {USA}}, pages 5547--5569.

\bibitem[{Du et~al.(2022{\natexlab{b}})Du, Qian, Liu, Ding, Qiu, Yang, and
  Tang}]{du-etal-2022-glm}
Zhengxiao Du, Yujie Qian, Xiao Liu, Ming Ding, Jiezhong Qiu, Zhilin Yang, and
  Jie Tang. 2022{\natexlab{b}}.
\newblock \href {https://doi.org/10.18653/v1/2022.acl-long.26} {{GLM}: General
  language model pretraining with autoregressive blank infilling}.
\newblock In \emph{Proceedings of the 60th Annual Meeting of the Association
  for Computational Linguistics (Volume 1: Long Papers)}, pages 320--335,
  Dublin, Ireland. Association for Computational Linguistics.

\bibitem[{Gokaslan et~al.(2019)Gokaslan, Pavlick, and
  Tellex}]{Gokaslan2019OpenWeb}
Aaron Gokaslan, Vanya Cohen~Ellie Pavlick, and Stefanie Tellex. 2019.
\newblock Openwebtext corpus.
\newblock \url{http://Skylion007.github.io/OpenWebTextCorpus}.

\bibitem[{Gu et~al.(2023)Gu, Zhu, Ye, Zhang, Xiong, Li, He, Jiang, Feng, and
  Xiao}]{gu2023domain}
Zhouhong Gu, Xiaoxuan Zhu, Haoning Ye, Lin Zhang, Zhuozhi Xiong, Zihan Li,
  Qianyu He, Sihang Jiang, Hongwei Feng, and Yanghua Xiao. 2023.
\newblock Domain mastery benchmark: An ever-updating benchmark for evaluating
  holistic domain knowledge of large language model--a preliminary release.
\newblock \emph{arXiv preprint arXiv:2304.11679}.

\bibitem[{Hendrycks et~al.(2021)Hendrycks, Burns, Basart, Zou, Mazeika, Song,
  and Steinhardt}]{Hendrycks-ICLR-2021-Measuring}
Dan Hendrycks, Collin Burns, Steven Basart, Andy Zou, Mantas Mazeika, Dawn
  Song, and Jacob Steinhardt. 2021.
\newblock Measuring massive multitask language understanding.
\newblock In \emph{{ICLR}}. OpenReview.net.

\bibitem[{Hoffmann et~al.(2022)Hoffmann, Borgeaud, Mensch, Buchatskaya, Cai,
  Rutherford, de~Las~Casas, Hendricks, Welbl, Clark, Hennigan, Noland,
  Millican, van~den Driessche, Damoc, Guy, Osindero, Simonyan, Elsen, Rae,
  Vinyals, and Sifre}]{Hoffmann-arxiv-2022-Training}
Jordan Hoffmann, Sebastian Borgeaud, Arthur Mensch, Elena Buchatskaya, Trevor
  Cai, Eliza Rutherford, Diego de~Las~Casas, Lisa~Anne Hendricks, Johannes
  Welbl, Aidan Clark, Tom Hennigan, Eric Noland, Katie Millican, George van~den
  Driessche, Bogdan Damoc, Aurelia Guy, Simon Osindero, Karen Simonyan, Erich
  Elsen, Jack~W. Rae, Oriol Vinyals, and Laurent Sifre. 2022.
\newblock Training compute-optimal large language models.
\newblock abs/2203.15556.

\bibitem[{Huang et~al.(2023)Huang, Dong, Wang, Hao, Singhal, Ma, Lv, Cui,
  Mohammed, Patra, Liu, Aggarwal, Chi, Bjorck, Chaudhary, Som, Song, and
  Wei}]{Huang-CoRR-2023}
Shaohan Huang, Li~Dong, Wenhui Wang, Yaru Hao, Saksham Singhal, Shuming Ma,
  Tengchao Lv, Lei Cui, Owais~Khan Mohammed, Barun Patra, Qiang Liu, Kriti
  Aggarwal, Zewen Chi, Johan Bjorck, Vishrav Chaudhary, Subhojit Som, Xia Song,
  and Furu Wei. 2023.
\newblock Language is not all you need: Aligning perception with language
  models.
\newblock \emph{CoRR}, abs/2302.14045.

\bibitem[{Khot et~al.(2020)Khot, Clark, Guerquin, Jansen, and
  Sabharwal}]{Khot-AAAI-2020-QASC}
Tushar Khot, Peter Clark, Michal Guerquin, Peter Jansen, and Ashish Sabharwal.
  2020.
\newblock {QASC:} {A} dataset for question answering via sentence composition.
\newblock In \emph{The Thirty-Fourth {AAAI} Conference on Artificial
  Intelligence, {AAAI} 2020, The Thirty-Second Innovative Applications of
  Artificial Intelligence Conference, {IAAI} 2020, The Tenth {AAAI} Symposium
  on Educational Advances in Artificial Intelligence, {EAAI} 2020, New York,
  NY, USA, February 7-12, 2020}, pages 8082--8090.

\bibitem[{Liu et~al.(2019{\natexlab{a}})Liu, He, Chen, and
  Gao}]{Liu-ACL-2019-Multi}
Xiaodong Liu, Pengcheng He, Weizhu Chen, and Jianfeng Gao. 2019{\natexlab{a}}.
\newblock Multi-task deep neural networks for natural language understanding.
\newblock In \emph{{ACL} {(1)}}, pages 4487--4496. Association for
  Computational Linguistics.

\bibitem[{Liu et~al.(2019{\natexlab{b}})Liu, Ott, Goyal, Du, Joshi, Chen, Levy,
  Lewis, Zettlemoyer, and Stoyanov}]{Liu-CoRR-2019-RoBERTa}
Yinhan Liu, Myle Ott, Naman Goyal, Jingfei Du, Mandar Joshi, Danqi Chen, Omer
  Levy, Mike Lewis, Luke Zettlemoyer, and Veselin Stoyanov. 2019{\natexlab{b}}.
\newblock Roberta: {A} robustly optimized {BERT} pretraining approach.
\newblock \emph{CoRR}, abs/1907.11692.

\bibitem[{Muennighoff et~al.(2022)Muennighoff, Wang, Sutawika, Roberts,
  Biderman, Scao, Bari, Shen, Yong, Schoelkopf, Tang, Radev, Aji, Almubarak,
  Albanie, Alyafeai, Webson, Raff, and
  Raffel}]{Muennighoff-2022-arxiv-Crosslingual}
Niklas Muennighoff, Thomas Wang, Lintang Sutawika, Adam Roberts, Stella
  Biderman, Teven~Le Scao, M.~Saiful Bari, Sheng Shen, Zheng~Xin Yong, Hailey
  Schoelkopf, Xiangru Tang, Dragomir Radev, Alham~Fikri Aji, Khalid Almubarak,
  Samuel Albanie, Zaid Alyafeai, Albert Webson, Edward Raff, and Colin Raffel.
  2022.
\newblock Crosslingual generalization through multitask finetuning.
\newblock \emph{CoRR}, abs/2211.01786.

\bibitem[{OpenAI(2023)}]{OpenAI-OpenAI-2023-GPT-4}
OpenAI. 2023.
\newblock Gpt-4 technical report.
\newblock \emph{OpenAI}.

\bibitem[{Ouyang et~al.(2022)Ouyang, Wu, Jiang, Almeida, Wainwright, Mishkin,
  Zhang, Agarwal, Slama, Ray, Schulman, Hilton, Kelton, Miller, Simens, Askell,
  Welinder, Christiano, Leike, and Lowe}]{Ouyang-arxiv-2022-Training}
Long Ouyang, Jeff Wu, Xu~Jiang, Diogo Almeida, Carroll~L. Wainwright, Pamela
  Mishkin, Chong Zhang, Sandhini Agarwal, Katarina Slama, Alex Ray, John
  Schulman, Jacob Hilton, Fraser Kelton, Luke Miller, Maddie Simens, Amanda
  Askell, Peter Welinder, Paul~F. Christiano, Jan Leike, and Ryan Lowe. 2022.
\newblock Training language models to follow instructions with human feedback.
\newblock \emph{CoRR}, abs/2203.02155.

\bibitem[{Paperno et~al.(2016)Paperno, Kruszewski, Lazaridou, Pham, Bernardi,
  Pezzelle, Baroni, Boleda, and Fern{\'{a}}ndez}]{Paperno-ACL-2016-LAMBADA}
Denis Paperno, Germ{\'{a}}n Kruszewski, Angeliki Lazaridou, Quan~Ngoc Pham,
  Raffaella Bernardi, Sandro Pezzelle, Marco Baroni, Gemma Boleda, and Raquel
  Fern{\'{a}}ndez. 2016.
\newblock The {LAMBADA} dataset: Word prediction requiring a broad discourse
  context.
\newblock In \emph{{ACL} {(1)}}. The Association for Computer Linguistics.

\bibitem[{Peng et~al.(2023)Peng, Li, He, Galley, and Gao}]{peng2023instruction}
Baolin Peng, Chunyuan Li, Pengcheng He, Michel Galley, and Jianfeng Gao. 2023.
\newblock Instruction tuning with gpt-4.
\newblock \emph{arXiv preprint arXiv:2304.03277}.

\bibitem[{Rae et~al.(2021)Rae, Borgeaud, Cai, Millican, Hoffmann, Song,
  Aslanides, Henderson, Ring, Young, Rutherford, Hennigan, Menick, Cassirer,
  Powell, van~den Driessche, Hendricks, Rauh, Huang, Glaese, Welbl, Dathathri,
  Huang, Uesato, Mellor, Higgins, Creswell, McAleese, Wu, Elsen, Jayakumar,
  Buchatskaya, Budden, Sutherland, Simonyan, Paganini, Sifre, Martens, Li,
  Kuncoro, Nematzadeh, Gribovskaya, Donato, Lazaridou, Mensch, Lespiau,
  Tsimpoukelli, Grigorev, Fritz, Sottiaux, Pajarskas, Pohlen, Gong, Toyama,
  de~Masson~d'Autume, Li, Terzi, Mikulik, Babuschkin, Clark, de~Las~Casas, Guy,
  Jones, Bradbury, Johnson, Hechtman, Weidinger, Gabriel, Isaac, Lockhart,
  Osindero, Rimell, Dyer, Vinyals, Ayoub, Stanway, Bennett, Hassabis,
  Kavukcuoglu, and Irving}]{Rae-arxiv-2021-Scaling}
Jack~W. Rae, Sebastian Borgeaud, Trevor Cai, Katie Millican, Jordan Hoffmann,
  H.~Francis Song, John Aslanides, Sarah Henderson, Roman Ring, Susannah Young,
  Eliza Rutherford, Tom Hennigan, Jacob Menick, Albin Cassirer, Richard Powell,
  George van~den Driessche, Lisa~Anne Hendricks, Maribeth Rauh, Po{-}Sen Huang,
  Amelia Glaese, Johannes Welbl, Sumanth Dathathri, Saffron Huang, Jonathan
  Uesato, John Mellor, Irina Higgins, Antonia Creswell, Nat McAleese, Amy Wu,
  Erich Elsen, Siddhant~M. Jayakumar, Elena Buchatskaya, David Budden, Esme
  Sutherland, Karen Simonyan, Michela Paganini, Laurent Sifre, Lena Martens,
  Xiang~Lorraine Li, Adhiguna Kuncoro, Aida Nematzadeh, Elena Gribovskaya,
  Domenic Donato, Angeliki Lazaridou, Arthur Mensch, Jean{-}Baptiste Lespiau,
  Maria Tsimpoukelli, Nikolai Grigorev, Doug Fritz, Thibault Sottiaux, Mantas
  Pajarskas, Toby Pohlen, Zhitao Gong, Daniel Toyama, Cyprien
  de~Masson~d'Autume, Yujia Li, Tayfun Terzi, Vladimir Mikulik, Igor
  Babuschkin, Aidan Clark, Diego de~Las~Casas, Aurelia Guy, Chris Jones, James
  Bradbury, Matthew~J. Johnson, Blake~A. Hechtman, Laura Weidinger, Iason
  Gabriel, William~S. Isaac, Edward Lockhart, Simon Osindero, Laura Rimell,
  Chris Dyer, Oriol Vinyals, Kareem Ayoub, Jeff Stanway, Lorrayne Bennett,
  Demis Hassabis, Koray Kavukcuoglu, and Geoffrey Irving. 2021.
\newblock Scaling language models: Methods, analysis {\&} insights from
  training gopher.
\newblock \emph{CoRR}, abs/2112.11446.

\bibitem[{Raffel et~al.(2020)Raffel, Shazeer, Roberts, Lee, Narang, Matena,
  Zhou, Li, and Liu}]{Raffel-JMLR-2020-Exploring}
Colin Raffel, Noam Shazeer, Adam Roberts, Katherine Lee, Sharan Narang, Michael
  Matena, Yanqi Zhou, Wei Li, and Peter~J. Liu. 2020.
\newblock Exploring the limits of transfer learning with a unified text-to-text
  transformer.
\newblock \emph{J. Mach. Learn. Res.}, pages 140:1--140:67.

\bibitem[{Rajpurkar et~al.(2016)Rajpurkar, Zhang, Lopyrev, and
  Liang}]{Rajpurkar-EMNLP-2016-SQuAD}
Pranav Rajpurkar, Jian Zhang, Konstantin Lopyrev, and Percy Liang. 2016.
\newblock Squad: 100, 000+ questions for machine comprehension of text.
\newblock In \emph{Proceedings of the 2016 Conference on Empirical Methods in
  Natural Language Processing, {EMNLP} 2016, Austin, Texas, USA, November 1-4,
  2016}, pages 2383--2392.

\bibitem[{Ren et~al.(2023)Ren, Zhou, Meng, Huang, Wang, Wang, Li, Zhang,
  Podolskiy, Arshinov, Bout, Piontkovskaya, Wei, Jiang, Su, Liu, and
  Yao}]{Ren-arXiv-2023-PanGusigma}
Xiaozhe Ren, Pingyi Zhou, Xinfan Meng, Xinjing Huang, Yadao Wang, Weichao Wang,
  Pengfei Li, Xiaoda Zhang, Alexander Podolskiy, Grigory Arshinov, Andrey Bout,
  Irina Piontkovskaya, Jiansheng Wei, Xin Jiang, Teng Su, Qun Liu, and Jun Yao.
  2023.
\newblock Pangu-{\(\Sigma\)}: Towards trillion parameter language model with
  sparse heterogeneous computing.
\newblock \emph{CoRR}, abs/2303.10845.

\bibitem[{Sanh et~al.(2022)Sanh, Webson, Raffel, Bach, Sutawika, Alyafeai,
  Chaffin, Stiegler, Raja, Dey, Bari, Xu, Thakker, Sharma, Szczechla, Kim,
  Chhablani, Nayak, Datta, Chang, Jiang, Wang, Manica, Shen, Yong, Pandey,
  Bawden, Wang, Neeraj, Rozen, Sharma, Santilli, F{\'{e}}vry, Fries, Teehan,
  Scao, Biderman, Gao, Wolf, and Rush}]{Sanh-ICLR-2022-Multitask}
Victor Sanh, Albert Webson, Colin Raffel, Stephen~H. Bach, Lintang Sutawika,
  Zaid Alyafeai, Antoine Chaffin, Arnaud Stiegler, Arun Raja, Manan Dey,
  M~Saiful Bari, Canwen Xu, Urmish Thakker, Shanya~Sharma Sharma, Eliza
  Szczechla, Taewoon Kim, Gunjan Chhablani, Nihal~V. Nayak, Debajyoti Datta,
  Jonathan Chang, Mike~Tian{-}Jian Jiang, Han Wang, Matteo Manica, Sheng Shen,
  Zheng~Xin Yong, Harshit Pandey, Rachel Bawden, Thomas Wang, Trishala Neeraj,
  Jos Rozen, Abheesht Sharma, Andrea Santilli, Thibault F{\'{e}}vry, Jason~Alan
  Fries, Ryan Teehan, Teven~Le Scao, Stella Biderman, Leo Gao, Thomas Wolf, and
  Alexander~M. Rush. 2022.
\newblock Multitask prompted training enables zero-shot task generalization.
\newblock In \emph{The Tenth International Conference on Learning
  Representations, {ICLR} 2022, Virtual Event, April 25-29, 2022}.
  OpenReview.net.

\bibitem[{Scao et~al.(2022)Scao, Fan, Akiki, Pavlick, Ilic, Hesslow,
  Castagn{\'{e}}, Luccioni, Yvon, Gall{\'{e}}, Tow, Rush, Biderman, Webson,
  Ammanamanchi, Wang, Sagot, Muennighoff, del Moral, Ruwase, Bawden, Bekman,
  McMillan{-}Major, Beltagy, Nguyen, Saulnier, Tan, Suarez, Sanh,
  Lauren{\c{c}}on, Jernite, Launay, Mitchell, Raffel, Gokaslan, Simhi, Soroa,
  Aji, Alfassy, Rogers, Nitzav, Xu, Mou, Emezue, Klamm, Leong, van Strien,
  Adelani, and et~al.}]{Scao-arxiv-2022-BLOOM}
Teven~Le Scao, Angela Fan, Christopher Akiki, Ellie Pavlick, Suzana Ilic,
  Daniel Hesslow, Roman Castagn{\'{e}}, Alexandra~Sasha Luccioni,
  Fran{\c{c}}ois Yvon, Matthias Gall{\'{e}}, Jonathan Tow, Alexander~M. Rush,
  Stella Biderman, Albert Webson, Pawan~Sasanka Ammanamanchi, Thomas Wang,
  Beno{\^{\i}}t Sagot, Niklas Muennighoff, Albert~Villanova del Moral, Olatunji
  Ruwase, Rachel Bawden, Stas Bekman, Angelina McMillan{-}Major, Iz~Beltagy,
  Huu Nguyen, Lucile Saulnier, Samson Tan, Pedro~Ortiz Suarez, Victor Sanh,
  Hugo Lauren{\c{c}}on, Yacine Jernite, Julien Launay, Margaret Mitchell, Colin
  Raffel, Aaron Gokaslan, Adi Simhi, Aitor Soroa, Alham~Fikri Aji, Amit
  Alfassy, Anna Rogers, Ariel~Kreisberg Nitzav, Canwen Xu, Chenghao Mou, Chris
  Emezue, Christopher Klamm, Colin Leong, Daniel van Strien, David~Ifeoluwa
  Adelani, and et~al. 2022.
\newblock {BLOOM:} {A} 176b-parameter open-access multilingual language model.
\newblock \emph{CoRR}, abs/2211.05100.

\bibitem[{Srivastava et~al.(2022)Srivastava, Rastogi, Rao, Shoeb, Abid, Fisch,
  Brown, Santoro, Gupta, Garriga{-}Alonso, Kluska, Lewkowycz, Agarwal, Power,
  Ray, Warstadt, Kocurek, Safaya, Tazarv, Xiang, Parrish, Nie, Hussain, Askell,
  Dsouza, Rahane, Iyer, Andreassen, Santilli, Stuhlm{\"{u}}ller, Dai, La,
  Lampinen, Zou, Jiang, Chen, Vuong, Gupta, Gottardi, Norelli, Venkatesh,
  Gholamidavoodi, Tabassum, Menezes, Kirubarajan, Mullokandov, Sabharwal,
  Herrick, Efrat, Erdem, Karakas, and et~al.}]{Srivastava-arxiv-2022-Beyond}
Aarohi Srivastava, Abhinav Rastogi, Abhishek Rao, Abu Awal~Md Shoeb, Abubakar
  Abid, Adam Fisch, Adam~R. Brown, Adam Santoro, Aditya Gupta, Adri{\`{a}}
  Garriga{-}Alonso, Agnieszka Kluska, Aitor Lewkowycz, Akshat Agarwal, Alethea
  Power, Alex Ray, Alex Warstadt, Alexander~W. Kocurek, Ali Safaya, Ali Tazarv,
  Alice Xiang, Alicia Parrish, Allen Nie, Aman Hussain, Amanda Askell, Amanda
  Dsouza, Ameet Rahane, Anantharaman~S. Iyer, Anders Andreassen, Andrea
  Santilli, Andreas Stuhlm{\"{u}}ller, Andrew~M. Dai, Andrew La, Andrew~K.
  Lampinen, Andy Zou, Angela Jiang, Angelica Chen, Anh Vuong, Animesh Gupta,
  Anna Gottardi, Antonio Norelli, Anu Venkatesh, Arash Gholamidavoodi, Arfa
  Tabassum, Arul Menezes, Arun Kirubarajan, Asher Mullokandov, Ashish
  Sabharwal, Austin Herrick, Avia Efrat, Aykut Erdem, Ayla Karakas, and et~al.
  2022.
\newblock Beyond the imitation game: Quantifying and extrapolating the
  capabilities of language models.
\newblock \emph{CoRR}, abs/2206.04615.

\bibitem[{Stiennon et~al.(2020)Stiennon, Ouyang, Wu, Ziegler, Lowe, Voss,
  Radford, Amodei, and Christiano}]{Stiennon-arxiv-2020-learning}
Nisan Stiennon, Long Ouyang, Jeff Wu, Daniel~M. Ziegler, Ryan Lowe, Chelsea
  Voss, Alec Radford, Dario Amodei, and Paul~F. Christiano. 2020.
\newblock Learning to summarize from human feedback.
\newblock \emph{CoRR}, abs/2009.01325.

\bibitem[{Sun et~al.(2021)Sun, Wang, Feng, Ding, Pang, Shang, Liu, Chen, Zhao,
  Lu, Liu, Wu, Gong, Liang, Shang, Sun, Liu, Ouyang, Yu, Tian, Wu, and
  Wang}]{Sun-arXiv-2021-ERNIE3.0}
Yu~Sun, Shuohuan Wang, Shikun Feng, Siyu Ding, Chao Pang, Junyuan Shang,
  Jiaxiang Liu, Xuyi Chen, Yanbin Zhao, Yuxiang Lu, Weixin Liu, Zhihua Wu,
  Weibao Gong, Jianzhong Liang, Zhizhou Shang, Peng Sun, Wei Liu, Xuan Ouyang,
  Dianhai Yu, Hao Tian, Hua Wu, and Haifeng Wang. 2021.
\newblock {ERNIE} 3.0: Large-scale knowledge enhanced pre-training for language
  understanding and generation.
\newblock \emph{CoRR}, abs/2107.02137.

\bibitem[{Taori et~al.(2023)Taori, Gulrajani, Zhang, Dubois, Li, Guestrin,
  Liang, and Hashimoto}]{alpaca}
Rohan Taori, Ishaan Gulrajani, Tianyi Zhang, Yann Dubois, Xuechen Li, Carlos
  Guestrin, Percy Liang, and Tatsunori~B. Hashimoto. 2023.
\newblock Stanford alpaca: An instruction-following llama model.
\newblock \url{https://github.com/tatsu-lab/stanford_alpaca}.

\bibitem[{Touvron et~al.(2023)Touvron, Lavril, Izacard, Martinet, Lachaux,
  Lacroix, Rozi{\`{e}}re, Goyal, Hambro, Azhar, Rodriguez, Joulin, Grave, and
  Lample}]{Touvron-arxiv-2023-LLaMA}
Hugo Touvron, Thibaut Lavril, Gautier Izacard, Xavier Martinet, Marie{-}Anne
  Lachaux, Timoth{\'{e}}e Lacroix, Baptiste Rozi{\`{e}}re, Naman Goyal, Eric
  Hambro, Faisal Azhar, Aur{\'{e}}lien Rodriguez, Armand Joulin, Edouard Grave,
  and Guillaume Lample. 2023.
\newblock Llama: Open and efficient foundation language models.
\newblock \emph{CoRR}.

\bibitem[{Wang et~al.(2019)Wang, Pruksachatkun, Nangia, Singh, Michael, Hill,
  Levy, and Bowman}]{Wang-NIPS-2019-SuperGLUE}
Alex Wang, Yada Pruksachatkun, Nikita Nangia, Amanpreet Singh, Julian Michael,
  Felix Hill, Omer Levy, and Samuel~R. Bowman. 2019.
\newblock Superglue: {A} stickier benchmark for general-purpose language
  understanding systems.
\newblock In \emph{NeurIPS}, pages 3261--3275.

\bibitem[{Wang et~al.(2018)Wang, Singh, Michael, Hill, Levy, and
  Bowman}]{Wang-EMNLP-2018-GLUE}
Alex Wang, Amanpreet Singh, Julian Michael, Felix Hill, Omer Levy, and
  Samuel~R. Bowman. 2018.
\newblock {GLUE:} {A} multi-task benchmark and analysis platform for natural
  language understanding.
\newblock In \emph{Proceedings of the Workshop: Analyzing and Interpreting
  Neural Networks for NLP, BlackboxNLP@EMNLP 2018, Brussels, Belgium, November
  1, 2018}, pages 353--355. Association for Computational Linguistics.

\bibitem[{Wang et~al.(2021)Wang, Sun, Xiang, Wu, Ding, Gong, Feng, Shang, Zhao,
  Pang, Liu, Chen, Lu, Liu, Wang, Bai, Chen, Zhao, Li, Sun, Yu, Ma, Tian, Wu,
  Wu, Zeng, Li, Gao, and Wang}]{Wang-arxiv-2021-ERNIE}
Shuohuan Wang, Yu~Sun, Yang Xiang, Zhihua Wu, Siyu Ding, Weibao Gong, Shikun
  Feng, Junyuan Shang, Yanbin Zhao, Chao Pang, Jiaxiang Liu, Xuyi Chen, Yuxiang
  Lu, Weixin Liu, Xi~Wang, Yangfan Bai, Qiuliang Chen, Li~Zhao, Shiyong Li,
  Peng Sun, Dianhai Yu, Yanjun Ma, Hao Tian, Hua Wu, Tian Wu, Wei Zeng, Ge~Li,
  Wen Gao, and Haifeng Wang. 2021.
\newblock {ERNIE} 3.0 titan: Exploring larger-scale knowledge enhanced
  pre-training for language understanding and generation.
\newblock \emph{CoRR}, abs/2112.12731.

\bibitem[{Wang et~al.(2022{\natexlab{a}})Wang, Kordi, Mishra, Liu, Smith,
  Khashabi, and Hajishirzi}]{Wang-arXiv-2022-Self}
Yizhong Wang, Yeganeh Kordi, Swaroop Mishra, Alisa Liu, Noah~A. Smith, Daniel
  Khashabi, and Hannaneh Hajishirzi. 2022{\natexlab{a}}.
\newblock Self-instruct: Aligning language model with self generated
  instructions.
\newblock \emph{CoRR}, abs/2212.10560.

\bibitem[{Wang et~al.(2022{\natexlab{b}})Wang, Mishra, Alipoormolabashi, Kordi,
  Mirzaei, Naik, Ashok, Dhanasekaran, Arunkumar, Stap, Pathak, Karamanolakis,
  Lai, Purohit, Mondal, Anderson, Kuznia, Doshi, Pal, Patel, Moradshahi,
  Parmar, Purohit, Varshney, Kaza, Verma, Puri, Karia, Doshi, Sampat, Mishra,
  A, Patro, Dixit, and Shen}]{Wang-EMNLP-2022-Super}
Yizhong Wang, Swaroop Mishra, Pegah Alipoormolabashi, Yeganeh Kordi, Amirreza
  Mirzaei, Atharva Naik, Arjun Ashok, Arut~Selvan Dhanasekaran, Anjana
  Arunkumar, David Stap, Eshaan Pathak, Giannis Karamanolakis, Haizhi~Gary Lai,
  Ishan Purohit, Ishani Mondal, Jacob Anderson, Kirby Kuznia, Krima Doshi,
  Kuntal~Kumar Pal, Maitreya Patel, Mehrad Moradshahi, Mihir Parmar, Mirali
  Purohit, Neeraj Varshney, Phani~Rohitha Kaza, Pulkit Verma, Ravsehaj~Singh
  Puri, Rushang Karia, Savan Doshi, Shailaja~Keyur Sampat, Siddhartha Mishra,
  Sujan~Reddy A, Sumanta Patro, Tanay Dixit, and Xudong Shen.
  2022{\natexlab{b}}.
\newblock Super-naturalinstructions: Generalization via declarative
  instructions on 1600+ {NLP} tasks.
\newblock In \emph{Proceedings of the 2022 Conference on Empirical Methods in
  Natural Language Processing, {EMNLP} 2022, Abu Dhabi, United Arab Emirates,
  December 7-11, 2022}, pages 5085--5109.

\bibitem[{Wei et~al.(2022)Wei, Bosma, Zhao, Guu, Yu, Lester, Du, Dai, and
  Le}]{Wei-ICLR-2022-Finetuned}
Jason Wei, Maarten Bosma, Vincent~Y. Zhao, Kelvin Guu, Adams~Wei Yu, Brian
  Lester, Nan Du, Andrew~M. Dai, and Quoc~V. Le. 2022.
\newblock Finetuned language models are zero-shot learners.
\newblock In \emph{The Tenth International Conference on Learning
  Representations, {ICLR} 2022, Virtual Event, April 25-29, 2022}.
  OpenReview.net.

\bibitem[{Wu et~al.(2021)Wu, Zhao, Yu, Zhang, Shen, Liu, Li, Zhu, Luo, Xu
  et~al.}]{Wu-arxiv-2021-Yuan}
Shaohua Wu, Xudong Zhao, Tong Yu, Rongguo Zhang, Chong Shen, Hongli Liu, Feng
  Li, Hong Zhu, Jiangang Luo, Liang Xu, et~al. 2021.
\newblock Yuan 1.0: Large-scale pre-trained language model in zero-shot and
  few-shot learning.
\newblock \emph{arXiv preprint arXiv:2110.04725}.

\bibitem[{Xie et~al.(2022)Xie, Raghunathan, Liang, and
  Ma}]{Michael-ICLR-2022-An}
Sang~Michael Xie, Aditi Raghunathan, Percy Liang, and Tengyu Ma. 2022.
\newblock An explanation of in-context learning as implicit bayesian inference.
\newblock In \emph{The Tenth International Conference on Learning
  Representations, {ICLR} 2022, Virtual Event, April 25-29, 2022}.

\bibitem[{Xu et~al.(2020)Xu, Hu, Zhang, Li, Cao, Li, Xu, Sun, Yu, Yu, Tian,
  Dong, Liu, Shi, Cui, Li, Zeng, Wang, Xie, Li, Patterson, Tian, Zhang, Zhou,
  Liu, Zhao, Zhao, Yue, Zhang, Yang, Richardson, and Lan}]{Xu-COLING-2020-CLUE}
Liang Xu, Hai Hu, Xuanwei Zhang, Lu~Li, Chenjie Cao, Yudong Li, Yechen Xu, Kai
  Sun, Dian Yu, Cong Yu, Yin Tian, Qianqian Dong, Weitang Liu, Bo~Shi, Yiming
  Cui, Junyi Li, Jun Zeng, Rongzhao Wang, Weijian Xie, Yanting Li, Yina
  Patterson, Zuoyu Tian, Yiwen Zhang, He~Zhou, Shaoweihua Liu, Zhe Zhao, Qipeng
  Zhao, Cong Yue, Xinrui Zhang, Zhengliang Yang, Kyle Richardson, and Zhenzhong
  Lan. 2020.
\newblock {CLUE:} {A} chinese language understanding evaluation benchmark.
\newblock In \emph{{COLING}}, pages 4762--4772. International Committee on
  Computational Linguistics.

\bibitem[{Xue et~al.(2021)Xue, Constant, Roberts, Kale, Al{-}Rfou, Siddhant,
  Barua, and Raffel}]{Xue-NAACL-2021-mT5}
Linting Xue, Noah Constant, Adam Roberts, Mihir Kale, Rami Al{-}Rfou, Aditya
  Siddhant, Aditya Barua, and Colin Raffel. 2021.
\newblock mt5: {A} massively multilingual pre-trained text-to-text transformer.
\newblock In \emph{Proceedings of the 2021 Conference of the North American
  Chapter of the Association for Computational Linguistics: Human Language
  Technologies, {NAACL-HLT} 2021, Online, June 6-11, 2021}, pages 483--498.

\bibitem[{Yunjie~Ji and Li(2023)}]{BELLE}
Yan Gong Yiping Peng Qiang Niu Baochang~Ma Yunjie~Ji, Yong~Deng and Xiangang
  Li. 2023.
\newblock Belle: Be everyone's large language model engine.
\newblock \url{https://github.com/LianjiaTech/BELLE}.

\bibitem[{Zellers et~al.(2019)Zellers, Holtzman, Rashkin, Bisk, Farhadi,
  Roesner, and Choi}]{Zellers-NeurIPS-2019-Defending}
Rowan Zellers, Ari Holtzman, Hannah Rashkin, Yonatan Bisk, Ali Farhadi,
  Franziska Roesner, and Yejin Choi. 2019.
\newblock Defending against neural fake news.
\newblock In \emph{Advances in Neural Information Processing Systems 32: Annual
  Conference on Neural Information Processing Systems 2019, NeurIPS 2019,
  December 8-14, 2019, Vancouver, BC, Canada}, pages 9051--9062.

\bibitem[{Zeng et~al.(2022)Zeng, Liu, Du, Wang, Lai, Ding, Yang, Xu, Zheng,
  Xia, Tam, Ma, Xue, Zhai, Chen, Zhang, Dong, and Tang}]{Zeng-arxiv-2022-GLM}
Aohan Zeng, Xiao Liu, Zhengxiao Du, Zihan Wang, Hanyu Lai, Ming Ding, Zhuoyi
  Yang, Yifan Xu, Wendi Zheng, Xiao Xia, Weng~Lam Tam, Zixuan Ma, Yufei Xue,
  Jidong Zhai, Wenguang Chen, Peng Zhang, Yuxiao Dong, and Jie Tang. 2022.
\newblock {GLM-130B:} an open bilingual pre-trained model.
\newblock abs/2210.02414.

\bibitem[{Zeng(2023)}]{zeng2023measuring}
Hui Zeng. 2023.
\newblock Measuring massive multitask chinese understanding.
\newblock \emph{arXiv preprint arXiv:2304.12986}.

\bibitem[{Zeng et~al.(2021)Zeng, Ren, Su, Wang, Liao, Wang, Jiang, Yang, Wang,
  Zhang, Li, Gong, Yao, Huang, Wang, Yu, Guo, Yu, Zhang, Wang, Tao, Yan, Yi,
  Peng, Jiang, Zhang, Deng, Zhang, Lin, Zhang, Zhang, Guo, Gu, Fan, Wang, Jin,
  Liu, and Tian}]{Zeng-arxiv-2021-PanGualpha}
Wei Zeng, Xiaozhe Ren, Teng Su, Hui Wang, Yi~Liao, Zhiwei Wang, Xin Jiang,
  ZhenZhang Yang, Kaisheng Wang, Xiaoda Zhang, Chen Li, Ziyan Gong, Yifan Yao,
  Xinjing Huang, Jun Wang, Jianfeng Yu, Qi~Guo, Yue Yu, Yan Zhang, Jin Wang,
  Hengtao Tao, Dasen Yan, Zexuan Yi, Fang Peng, Fangqing Jiang, Han Zhang,
  Lingfeng Deng, Yehong Zhang, Zhe Lin, Chao Zhang, Shaojie Zhang, Mingyue Guo,
  Shanzhi Gu, Gaojun Fan, Yaowei Wang, Xuefeng Jin, Qun Liu, and Yonghong Tian.
  2021.
\newblock Pangu-{\(\alpha\)}: Large-scale autoregressive pretrained chinese
  language models with auto-parallel computation.
\newblock \emph{CoRR}, abs/2104.12369.

\bibitem[{Zhang et~al.(2022)Zhang, Roller, Goyal, Artetxe, Chen, Chen, Dewan,
  Diab, Li, Lin, Mihaylov, Ott, Shleifer, Shuster, Simig, Koura, Sridhar, Wang,
  and Zettlemoyer}]{Zhang-arxiv-2022-OPT}
Susan Zhang, Stephen Roller, Naman Goyal, Mikel Artetxe, Moya Chen, Shuohui
  Chen, Christopher Dewan, Mona~T. Diab, Xian Li, Xi~Victoria Lin, Todor
  Mihaylov, Myle Ott, Sam Shleifer, Kurt Shuster, Daniel Simig, Punit~Singh
  Koura, Anjali Sridhar, Tianlu Wang, and Luke Zettlemoyer. 2022.
\newblock {OPT:} open pre-trained transformer language models.
\newblock \emph{CoRR}, abs/2205.01068.

\bibitem[{Zhang et~al.(2021)Zhang, Gu, Han, Chen, Xiao, Sun, Yao, Qi, Guan, Ke,
  Cai, Zeng, Tan, Liu, Huang, Han, Liu, Zhu, and Sun}]{Zhang-arXiv-2021-CPM-2}
Zhengyan Zhang, Yuxian Gu, Xu~Han, Shengqi Chen, Chaojun Xiao, Zhenbo Sun, Yuan
  Yao, Fanchao Qi, Jian Guan, Pei Ke, Yanzheng Cai, Guoyang Zeng, Zhixing Tan,
  Zhiyuan Liu, Minlie Huang, Wentao Han, Yang Liu, Xiaoyan Zhu, and Maosong
  Sun. 2021.
\newblock {CPM-2:} large-scale cost-effective pre-trained language models.
\newblock \emph{CoRR}, abs/2106.10715.

\bibitem[{Zhao et~al.(2023)Zhao, Zhou, Li, Tang, Wang, Hou, Min, Zhang, Zhang,
  Dong, Du, Yang, Chen, Chen, Jiang, Ren, Li, Tang, Liu, Liu, Nie, and
  Wen}]{LLMSurvey}
Wayne~Xin Zhao, Kun Zhou, Junyi Li, Tianyi Tang, Xiaolei Wang, Yupeng Hou,
  Yingqian Min, Beichen Zhang, Junjie Zhang, Zican Dong, Yifan Du, Chen Yang,
  Yushuo Chen, Zhipeng Chen, Jinhao Jiang, Ruiyang Ren, Yifan Li, Xinyu Tang,
  Zikang Liu, Peiyu Liu, Jian-Yun Nie, and Ji-Rong Wen. 2023.
\newblock \href {http://arxiv.org/abs/2303.18223} {A survey of large language
  models}.
\newblock \emph{arXiv preprint arXiv:2303.18223}.

\bibitem[{Zhong et~al.(2023)Zhong, Cui, Guo, Liang, Lu, Wang, Saied, Chen, and
  Duan}]{zhong2023agieval}
Wanjun Zhong, Ruixiang Cui, Yiduo Guo, Yaobo Liang, Shuai Lu, Yanlin Wang, Amin
  Saied, Weizhu Chen, and Nan Duan. 2023.
\newblock Agieval: A human-centric benchmark for evaluating foundation models.
\newblock \emph{arXiv preprint arXiv:2304.06364}.

\bibitem[{Zhou et~al.(2023)Zhou, Li, Li, Yu, Liu, Wang, Zhang, Ji, Yan, He,
  Peng, Li, Wu, Liu, Xie, Xiong, Pei, Yu, and Sun}]{Zhou-arxiv-2023-A}
Ce~Zhou, Qian Li, Chen Li, Jun Yu, Yixin Liu, Guangjing Wang, Kai Zhang, Cheng
  Ji, Qiben Yan, Lifang He, Hao Peng, Jianxin Li, Jia Wu, Ziwei Liu, Pengtao
  Xie, Caiming Xiong, Jian Pei, Philip~S. Yu, and Lichao Sun. 2023.
\newblock A comprehensive survey on pretrained foundation models: {A} history
  from {BERT} to chatgpt.
\newblock \emph{CoRR}, abs/2302.09419.

\bibitem[{Zhu et~al.(2015)Zhu, Kiros, Zemel, Salakhutdinov, Urtasun, Torralba,
  and Fidler}]{Zhu-ICCV-2015-Aligning}
Yukun Zhu, Ryan Kiros, Richard~S. Zemel, Ruslan Salakhutdinov, Raquel Urtasun,
  Antonio Torralba, and Sanja Fidler. 2015.
\newblock Aligning books and movies: Towards story-like visual explanations by
  watching movies and reading books.
\newblock In \emph{2015 {IEEE} International Conference on Computer Vision,
  {ICCV} 2015, Santiago, Chile, December 7-13, 2015}, pages 19--27. {IEEE}
  Computer Society.

\end{thebibliography}
\bibliographystyle{acl_natbib}


\begin{center}  
\onecolumn
\begin{longtable}{lcc} 
\hline
\textbf{Tasks}&\textbf{Subjects} &\textbf{Education System} \\
\hline
\text{Chinese} & \text{Arts \& Humanities}&\text{Primary school}\\
\text{Math} & \text{Natural Sciences}&\text{Primary school}\\
\hline
\text{Chinese} & \text{Arts \& Humanities}&\text{Junior high school}\\
\text{History} & \text{Arts \& Humanities}&\text{Junior high school}\\
\text{Politics} & \text{Social Sciences}&\text{Junior high school}\\
\text{Math} & \text{Natural Sciences}&\text{Junior high school}\\
\text{Physics} & \text{Natural Sciences}&\text{Junior high school}\\
\text{Biology} & \text{Natural Sciences}&\text{Junior high school}\\
\text{Chemistry} & \text{Natural Sciences}&\text{Junior high school}\\
\text{Geography} & \text{Natural Sciences}&\text{Junior high school}\\
\hline
\text{Chinese} & \text{Arts \& Humanities}&\text{High school}\\
\text{History} & \text{Arts \& Humanities}&\text{High school}\\
\text{Politics} & \text{Social Sciences}&\text{High school}\\
\text{Math} & \text{Natural Sciences}&\text{High school}\\
\text{Physics} & \text{Natural Sciences}&\text{High school}\\
\text{Biology} & \text{Natural Sciences}&\text{High school}\\
\text{Chemistry} & \text{Natural Sciences}&\text{High school}\\
\text{Geography} & \text{Natural Sciences}&\text{High school}\\
\hline
\text{Modern History} & \text{Arts \& Humanities}&\text{College}\\
\text{History Foundation} & \text{Arts \& Humanities}&\text{College}\\
\text{Modern World History} & \text{Arts \& Humanities}&\text{College}\\
\text{Chinese Constitutional Law} & \text{Social Sciences}&\text{College}\\
\text{History of Chinese Education} & \text{Social Sciences}&\text{College}\\
\text{History of the Chinese Legal System} & \text{Social Sciences}&\text{College}\\
\text{Developmental and Educational Psychology} & \text{Social Sciences}&\text{College}\\
\text{History of Foreign Education} & \text{Social Sciences}&\text{College}\\
\text{Experimental Psychology} & \text{Social Sciences}&\text{College}\\
\text{Introduction to Psychology} & \text{Social Sciences}&\text{College}\\
\text{Moral Cultivation} & \text{Social Sciences}&\text{College}\\
\text{Psychology of Teaching} & \text{Social Sciences}&\text{College}\\
\text{Principles of Pedagogy} & \text{Social Sciences}&\text{College}\\
\text{Educational Research Methods} & \text{Social Sciences}&\text{College}\\
\text{Current Affairs and Politics} & \text{Social Sciences}&\text{College}\\
\text{Introduction to Mao Tsetung Thoughts} & \text{Social Sciences}&\text{College}\\
\text{Civil Law} & \text{Social Sciences}&\text{College}\\
\text{Jurisprudence} & \text{Social Sciences}&\text{College}\\
\text{Sociology} & \text{Social Sciences}&\text{College}\\
\text{Basic Principle of Marxism} & \text{Social Sciences}&\text{College}\\
\text{Criminal Jurisprudence} & \text{Social Sciences}&\text{College}\\
\text{Outline of Chinese Modern History} & \text{Social Sciences}&\text{College}\\
\text{Humanistic Medicine} & \text{Natural Sciences}&\text{College}\\
\text{Internal Medicine} & \text{Natural Sciences}&\text{College}\\
\text{Animal Physiology} & \text{Natural Sciences}&\text{College}\\
\text{Surgical Sciences} & \text{Natural Sciences}&\text{College}\\
\text{Operating Systems} & \text{Natural Sciences}&\text{College}\\
\text{Data Structures} & \text{Natural Sciences}&\text{College}\\
\text{Probability Theory} & \text{Natural Sciences}&\text{College}\\
\text{Biochemistry} & \text{Natural Sciences}&\text{College}\\
\text{Biochemistry and Pathology} & \text{Natural Sciences}&\text{College}\\
\text{Physiology} & \text{Natural Sciences}&\text{College}\\
\text{Principles of Computer Composition} & \text{Natural Sciences}&\text{College}\\
\text{Computer Networks} & \text{Natural Sciences}&\text{College}\\
\text{Advanced Mathematics} & \text{Natural Sciences}&\text{College}\\
\text{Linear Algebra} & \text{Natural Sciences}&\text{College}\\
\text{Stomatology} & \text{Natural Sciences}&\text{College}\\
\text{Anthropotomy} & \text{Natural Sciences}&\text{College}\\
\text{Pharmacology} & \text{Natural Sciences}&\text{College}\\
\text{Immunology} & \text{Natural Sciences}&\text{College}\\
\text{Management} & \text{Natural Sciences}&\text{College}\\
\text{Economics} & \text{Natural Sciences}&\text{College}\\
\hline
\text{Film} & \text{Arts \& Humanities}&\text{Other}\\
\text{Music} & \text{Arts \& Humanities}&\text{Other}\\
\text{Dance} & \text{Arts \& Humanities}&\text{Other}\\
\text{Fine Arts} & \text{Arts \& Humanities}&\text{Other}\\
\text{Computer Fundamentals} & \text{Natural Sciences}&\text{Other}\\
\text{Computer Programming Language} & \text{Natural Sciences}&\text{Other}\\
\text{Chinese Medicine} & \text{Other}&\text{Other}\\
\text{Ancient Chinese Language} & \text{Other}&\text{Other}\\
\text{Novels} & \text{Other}&\text{Other}\\
\text{Religion} & \text{Other}&\text{Other}\\
\text{Chinese Civil Service Examination} & \text{Other}&\text{Other}\\
\hline
\caption{Summary of all 71 tasks.} \label{summary} \
\end{longtable}  
\end{center}

\appendix
\section{All Subjects}
\label{sec:subjects}
See Table~\ref{summary} for all 71 tasks.

\end{document}